%% file: nips_2018.tex
\documentclass{article}

\usepackage[preprint]{nips_2018}

\usepackage[utf8]{inputenc} 
\usepackage[T1]{fontenc}    
\usepackage{hyperref}       
\usepackage{url}            
\usepackage{booktabs}       
\usepackage{amsfonts}       
\usepackage{nicefrac}       
\usepackage{microtype}      
\usepackage{algorithm} 
\usepackage{algpseudocode} 

\usepackage{caption}
\usepackage{subfig}

\newlength\myindent
\usepackage{mlmacros}
\usepackage{amsthm}
\usepackage{amsmath} 
\usepackage{amssymb}
\usepackage{cancel}
\usepackage{cleveref}

\probdists{p,q}
\probdists[policy,source,transition,plan,initial]{\pi,\omega,p,q,\rho}
\probdists[optimalpolicy,optimalsource]{\pi^{\star},\omega^{\star}}
\probdists[recog,recogmeas,recogtrans]{q,q_{\text{meas}},q_{\text{trans}}}
\MkProbDist{empirical}{\hat{p}}
\variables[obs,innov,state,action]{x,e,s,a}

\newcommand{\succstate}{{\bstate^\prime\!}}
\newcommand{\nowstate}{\bstate}
\newcommand{\nowaction}{\baction}

\newcommand{\margtrans}{\p{\succstate}{\nowstate}}
\newcommand{\jointtrans}{\p{\succstate, \nowaction}{\nowstate}}
\newcommand{\sourcetrans}{\source{\nowaction}{\nowstate}}

\newcommand{\emp}{\mathcal{E}}

\DeclareDocumentCommand\mi{oG{}G{}}{\ensuremath{\operatorname{\mathcal{I}}\IfNoValueOrEmptyTF{#1}{}{_{#1}}\IfNoValueOrEmptyTF{#2}{}{\mleft(#2 \IfNoValueOrEmptyTF{#3}{}{\mid #3}\mright)}}}

\DeclareDocumentCommand\mibound{oG{}G{}}{\ensuremath{\operatorname{\mathcal{\hat{I}}}\IfNoValueOrEmptyTF{#1}{}{_{#1}}\IfNoValueOrEmptyTF{#2}{}{\mleft(#2 \IfNoValueOrEmptyTF{#3}{}{\mid #3}\mright)}}}

\usepackage{tikz}

\title{Empowerment-driven Exploration using Mutual Information Estimation}

\author{
  Navneet Madhu Kumar\thanks{Code is available at https://github.com/navneet-nmk/pytorch-rl} \\
  Department of Computer Science\\
  Technical University of Munich\\
  Germany \\
  \texttt{navneet.nmk@gmail.com} \\
}

\begin{document}

\maketitle

\begin{abstract}
  Exploration is a difficult challenge in reinforcement learning and is of prime importance in sparse reward environments. However, many of the state of the art deep reinforcement learning algorithms, that reply on epsilon-greedy exploration, fail on these environments. In such environments, empowerment can serve as an intrinsic reward signal to enable the agent to explore by maximizing the influence it has over the near future. We formulate empowerment as the channel capacity between actions and states and is calculated by estimating the mutual information between the actions and the following states. The mutual information is estimated using a Mutual Information Neural Estimator and a forward dynamics model. We demonstrate that an empowerment driven agent is able to improve significantly the score of a baseline DQN agent on the game of Montezuma's Revenge.
\end{abstract}
\section{Introduction}

Reinforcement learning (RL) tackles sequential decision making problems by formulating them as
tasks where an agent must learn how to act optimally through trial and error interactions with the
environment.  The goal is to maximize the sum of the numerical reward signal
observed at each time step. 
These rewards are usually provided to the agent by the environment, 
either continuously or sparsely.
Here we focus on the problem of exploration in RL, which aims to reduce 
the number of interactions an agent needs in order to learn to perform well.

The most common approach to exploration in absence of any
knowledge about the environment is to perform random
actions.  As knowledge is gained, the agent can use it to
attempt to increase its performance by taking greedy actions,
while retaining some chance to choose random actions to
further explore the environment (epsilon-greedy exploration).
However, if rewards are sparse or are not sufficiently in-
formative to allow performance improvements,
epsilon-greedy fails  to  explore  sufficiently  far.    
Several  methods  have been described that bias the agent’s actions towards novelty
mostly by using optimistic initialization or curiosity signals.

In particular,we opt for empowerment \cite{Empowerment}, an information-theoretic formulation of
the agent’s influence on the near future. The value of a state, its empowerment value, is given by the maximum
mutual information between a control input and the successor state the agent could achieve.

Mutual Information is known to be very hard to calculate. Thankfully,  recent advances in
neural estimation \cite{MINE} enable effective computation of mutual information between high dimensional input / output pairs of
deep neural networks, and in this work we leverage these techniques to calculate the empowerment of a state. We then use this quantity as an intrinsic reward signal to train the DQN on Montezuma's Revenge.

\input{empowerment}

\input{experiment}

\input{conclusion}

\bibliography{refs}

\end{document}

%% file: empowerment.tex
\section{Empowerment-Driven Exploration}
\label{gen_inst}

Our agent is composed of two networks: a reward generator that outputs a empowerment-driven intrinsic reward signal
and  a  policy  that  outputs  a  sequence  of  actions  to  maximize that reward signal. In addition to intrinsic rewards,
the agent optionally may also receive some extrinsic reward
from  the  environment.  Let  the  intrinsic  curiosity  reward
generated by the agent at time t be rit and the extrinsic reward be ret. The policy sub-system is trained to maximize
the sum of these two rewards rt=rit+ret, with ret mostly(if not always) zero.

\subsection{Empowerment}
The Empowerment value for a state $\bstate$ is defined as the channel capacity between the action $\baction$ and the following state $\succstate$
\citet{Empowerment}, 
\eq{
    \emp(\nowstate) = \max_{\policy}\, \mi{\succstate, \baction}{\nowstate}. \numberthis \label{eq:empowerment}
}

Where I is the mutual information. $\policy$ is the empowerment maximizing policy. Empowerment, therefore, is the maximum information an agent can transfer to it's environment by changing the next states through it's actions. 

This mutual information in \cref{eq:empowerment} can further be represented in the form of the KL Divergence,
\eq{
    \mi{\succstate, \nowaction}{\nowstate} 
        =& \kl{\jointtrans}{\margtrans\,\source{\nowaction}{\nowstate}} \\
        =& \iint\!  \jointtrans \ln \frac{\jointtrans}{\margtrans \,\sourcetrans}\dint{\succstate} \dint{\nowaction},
}

To compute the mutual information, we will be using the formulation as explained in \cite{MINE}.

\subsection{Mutual Information Neural Estimation}
The Mutual Information Neural Estimation (MINE, \cite{MINE}) learns a neural estimate of the mutual information of continuous variables, is strongly consistent and can be used to learn the empowerment value of a state by using a forward dynamics model, $\p{\bstate_{t+1}}{\bstate_t, \baction_t}$ to get the samples from the marginal distribution of $\p{\bstate_{t+1}}{\bstate_t}$, and the policy, $\policy{\baction_t}{\bstate_t}$. 

Following, \cite{MINE}, we train a discriminator (a classifier) to distinguish between samples coming from the joint, $\jointtrans$, and the marginal distributions, $\p{\bstate_{t+1}}{\bstate_t}$ and $\policy{\baction_t}{\bstate_t}$. MINE relies on a lower-bound to the mutual information based on the \cite{DonskerVaradhan}, 

\eq{
	\mi{\succstate, \nowaction}{\nowstate}
    =& \kl{\jointtrans}{\margtrans\,\source{\nowaction}{\nowstate}} \\
    \geq&  \expc[\jointtrans][T_{\omega}] - \log\expc[\margtrans \policy{\nowaction}{\nowstate}][e^{T_{\omega}}]
}

We use this estimate of the mutual information (or the empowerment for the state, $\bstate$) as the intrinsic reward to train the policy. The policy is, therefore, encouraged to predict actions which provide the maximum mutual information.

One thing to note is that the distribution, $\p{\bstate_{t+1}}{\bstate_t}$, is calculated using the forward dynamics model which will be discussed in the following section. 

\subsection{Forward dynamics model}
The forward dynamics model, $\p{\bstate_{t+1}}{\bstate_t, \baction_t}$, is used to sample from the marginal distribution, $\p{\bstate_{t+1}}{\bstate_t}$ by marginalizing out the actions,
\begin{align*}
	\margtrans = \int \policy{\nowaction}{\nowstate} \, \jointtrans \dint{\nowaction}.
\end{align*}
 
The dynamics model is trained simultaneously with the statistics network, $T$ and the policy, $\policy{\nowaction}{\nowstate}$.
Predicting in the raw pixel space is not ideal since it is hard to predict pixels directly. So we use the random feature space as used in \cite{pathak18largescale}, to train the forward dynamics model as well as the policy since it was shown that the random feature space was sufficient for representing Atari game frames.

\begin{algorithm*}[t!]
		\caption{Joint training of policy $\policy$, statistics $T$ and forward dynamics model $\p{\bstate_{t+1}}{\bstate_t, \baction_t}$}
	\begin{algorithmic}
		\Require cumulation horizon $N$, initializations for $\policy$, $T$, $f$ and $e$
		\Repeat 
			\For{$m=1:M$}
            	\State {Sample a batch $b$ of transitions from a replay buffer $B$}
				\For{each transition ($\bstate_t$, $\baction_t$, $\bstate_{t+1}$, $r_{e}$) }
                	\State Update f to reduce $\sum_t[e(\bstate_{t+1}) - f(e(\bstate_t, \baction_t))]_{2}^2$
                    
					\State Update T to increase the mutual information
					\eq{
					\expc[\jointtrans][T_{\omega}] - \log\expc[\margtrans \policy{\nowaction}{\nowstate}][e^{T_{\omega}}] 
					}
					\State Update the reward function for the transition
					$r$ = $r_{e}$ + \mi{\succstate,\nowaction}{\nowstate}
				\EndFor
				\State Update the agent using the bellman update
				$y$ = $r$ + $\gamma*\max_{a} Q(\bstate_t, \baction_t)$
				
			\EndFor
		\Until{convergence}

	\end{algorithmic}

	\label{alg:training}
\end{algorithm*}

%% file: experiment.tex
\section{Experimental Setup}

\subsection{Agent}
The proposed empowerment driven DQN agent is composed of a policy network, the DQN, and the intrinsic reward generator that is composed of the statistics network $T$ and the forward dynamics model, $\p{\bstate_{t+1}}{\bstate_t, \baction_t}$.  The implementation is using Pytorch \cite{paszke2017automatic}. Inputs are a stack of four 84 x 84 gray scale frames. All the observations are then encoded to a 64 dimensional encoding using a random convolutional encoder. 
All other networks share the same network architecture but use separate artificial neural networks. We used double Q-learning with target network updates every 2000 steps and an experience replay of buffer capacity 1000000 steps. Training of the models starts at 1000 steps and follows Algorithm 1.  All 3 networks are trained on batch size of 64, using Adam optimizers with learning rate of 1e-2, 1e-3 and 1e-4 for the forward dynamics, statistics and policy network respectively.
\\
The DQN Agent predicts 18 Q-values corresponding to the 18 actions in ALE, uses relu activations and a discount factor of 0.99.
The extrinsic reward is clipped between [-1, 1] and the gradients of the temporal difference loss is clamped between [-1, 1].
The value used for $\beta$ is 0.1.  

\subsection{Results}

The DQN Agent trained using the empowerment intrinsic motivation is able to consistently exit the room one and gather the rewards whereas the agent trained on the reward signal of the game fails to receive any reward.

\begin{figure}[!htb]
  \includegraphics[width=\linewidth]{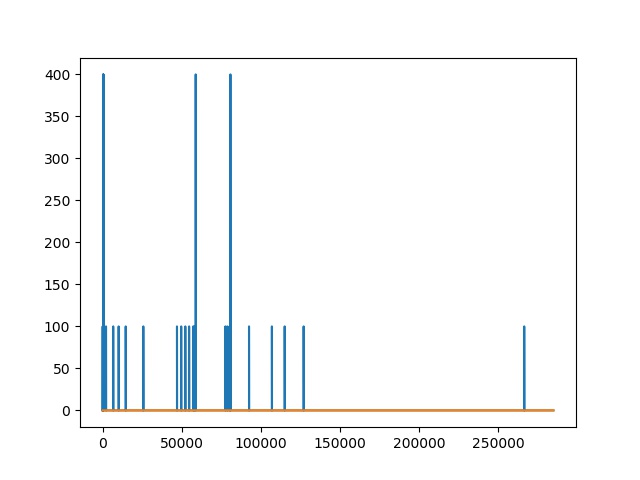}
  \caption{Rewards in Montezuma's Revenge}
  \label{fig:results}
\end{figure}

Owing to computational limits, the size of the environment encoding was limited to 64 which could prove to be insufficient for the Atari Environment. This is a parameter which needs to be investigated further in future work.

%% file: conclusion.tex
\section{Conclusion and future work}
\label{sec:conclusion_and_future_work}
The experiments show that using empowerment, calculated using mutual information neural estimation, as an intrinsic motivator can help an agent to consistently achieve rewards.

Compared to an agent which just receives the external reward signal from the game, the empowerment driven agent is able to consistently achieve the rewards in the first level of Montezuma's revenge and enter the second room.  

Using empowerment as an intrinsic motivator is a direction which has also been worked upon by previous research work but in this work we have the following two advantages. First, empowerment is calculated and maximized using stochastic gradient descent .
Second, in this work we just use the mutual information as the intrinsic reward and simply update the agent using Q-Learning. No new source policy distributions are introduced making the algorithm easy to implement.

Future work includes the testing of the method on the entire Atari game suite as well as increasing the model sizes and embedding sizes.